\documentclass[conference]{IEEEtran}
\IEEEoverridecommandlockouts
\usepackage{cite}
\usepackage{amsmath,amssymb,amsfonts}
\usepackage{algorithmic}
\usepackage{graphicx}
\usepackage{textcomp}
\usepackage{xcolor}
\def\BibTeX{{\rm B\kern-.05em{\sc i\kern-.025em b}\kern-.08em
    T\kern-.1667em\lower.7ex\hbox{E}\kern-.125emX}}
\begin{document}

\title{Quantitative Ink Analysis: Estimating the Number of Inks in Documents through Hyperspectral Imaging\\
}

\author{\IEEEauthorblockN{Aneeqa Abrar}
\IEEEauthorblockA{\textit{Department of Space Science} \\
\textit{Institute of Space Technology}\\
Islamabad, Pakistan \\
aneeqa.abrar@grel.ist.edu.pk}
\and
\IEEEauthorblockN{Hamza Iqbal}
\IEEEauthorblockA{\textit{Deparment of Electrical Engineering} \\
\textit{Institute of Space Technology}\\
Islamabad, Pakistan\\
hamzaiqbal.real@gmail.com }
}
\maketitle

\begin{abstract}
In the field of document forensics, ink analysis plays a crucial role in determining the authenticity of legal and historic documents and detecting forgery. Visual examination alone is insufficient for distinguishing visually similar inks, necessitating the use of advanced scientific techniques. This paper proposes an ink analysis technique based on hyperspectral imaging, which enables the examination of documents in hundreds of narrowly spaced spectral bands, revealing hidden details. The main objective of this study is to identify the number of distinct inks used in a document. Three clustering algorithms, namely k-means, Agglomerative, and c-means, are employed to estimate the number of inks present. The methodology involves data extraction, ink pixel segmentation, and ink number determination. The results demonstrate the effectiveness of the proposed technique in identifying ink clusters and distinguishing between different inks. The analysis of a hyperspectral cube dataset reveals variations in spectral reflectance across different bands and distinct spectral responses among the 12 lines, indicating the presence of multiple inks. The clustering algorithms successfully identify ink clusters, with k-means clustering showing superior classification performance. These findings contribute to the development of reliable methodologies for ink analysis using hyperspectral imaging, enhancing the accuracy and precision of forensic document examination.
\end{abstract}


\section{Introduction}
The ink analysis in historic documents helps in establishing the authenticity of these documents helping in identifying forgery, fraud, backdating and ink age. While the human eye possesses the ability to distinguish a limited electromagnetic spectrum and perceive various colors, it fails to discriminate visually similar inks with distinct spectral responses\cite{abbas2017towards}. Consequently, relying solely on visual examination is insufficient to differentiate between such inks, necessitating the utilization of advanced scientific techniques.

Over the years, two primary approaches for ink analysis techniques have emerged: destructive and non-destructive examination methods. Destructive methods, such as Thin Layer Chromatography (TLC)\cite{khan2013hyperspectral}, have traditionally been employed in forensic investigations to separate ink constituents. TLC exploits the unique chemical compositions and characteristic reactivity of inks used throughout historic documents to detect ink mismatches. However, this method has several drawbacks, including irreparable damage to the document, time-consuming procedures, sensitivity to temperature fluctuations, and the qualitative nature of the analysis. Furthermore, the quantification of TLC results requires meticulous considerations.

To address these limitations, Easton et al. \cite{easton2003multispectral} introduced hyperspectral imaging (HSI), an innovative and non-destructive technique capable of examining documents in hundreds of narrowly spaced spectral bands, thereby unveiling hidden details. Over time, increasingly sophisticated hyperspectral imaging systems have been developed to detect forgery and facilitate quantitative ink mismatch analysis.

In this paper, an ink analysis technique based on hyperspectral imaging is proposed for ink mismatch detection. Our main focus was to identify the number of inks used in the document. We used three different clustering algorithms, namely k-means, Agglomerative and C-means, to estimate the number of inks present in the image. The assumptions made for this paper were that different inks exhibit different spectral responses. However, no assumptions were made regarding the total number of inks present in the document.

This paper proposes an ink analysis technique based on hyperspectral imaging for the purpose of detecting ink mismatches. The primary objective was to identify the number of distinct inks utilized in a given document. To accomplish this, we used three distinct unsupervised clustering algorithms, namely k-means, Agglomerative, and c-means, to estimate the number of inks present within the hyperspectral image. It is important to note that while we assumed that different inks exhibit unique spectral responses, we made no assumptions regarding the total number of inks present in the document. This study also aims to aid the potential of hyperspectral imaging to enhance the scientific precision of ink analysis and contribute to the development of reliable methodologies for detecting ink mismatches in documents thereby safeguarding the authenticity and credibility of legal records and historical artifacts.

\section{Literature Review}
Hyperspectral imaging has emerged as a powerful tool in various fields, including remote sensing, material identification, and forensic document analysis. Recent advancements in sensor technology have enabled the capture of detailed spatial, spectral, and temporal data. This literature review aims to provide an overview of current trends and research in hyperspectral imaging, with a specific focus on its application in forensic document analysis.

In their research, Brauns and Dyer \cite{doi:10.1366/000370206778062093} devised a non-destructive hyperspectral imaging system to identify forgeries in potentially fraudulent documents. To simulate altered documents, they prepared written materials using blue, black, and red inks, and subsequently introduced modifications using a different ink of the same color. Padaon et al. \cite{padoan2008quantitative}, on the other hand, enhanced the hyperspectral imaging system specifically tailored for analyzing historical documents stored in archives. Their innovation involved the utilization of a narrow-band tunable light source, which effectively mitigated the risk of document damage caused by excessive heat generated by a powerful broadband white light source.

Khan et al. \cite{khan2018modern} discuss the progress in sensor technology and highlight a wide range of applications, including food quality assessment, medical surgery, forensic document examination, defense, precision agriculture, water resource management, artwork mapping, and deep learning-assisted forgery detection. One specific area of application within forensic document analysis is ink mismatch detection \cite{khan2013hyperspectral}, where the focus is on the effectiveness of hyperspectral imaging in differentiating visually similar inks. Khan et al. \cite{khan2013hyperspectral} demonstrate the limitations of traditional RGB scans and emphasize the superior ink discrimination achieved through hyperspectral imaging. This study highlights the potential of hyperspectral imaging in questioned document examination and encourages further research in this area.

Abbas et al. \cite{abbas2017towards} propose an automated ink mismatch detection approach using hyperspectral unmixing. The authors address the issue of disproportionate ink mismatch detection by identifying spectral signatures and proportions of inks. Automated forgery detection is another important area within forensic document analysis \cite{khan2018automated}, which utilizes Fuzzy C-Means Clustering for ink mismatch detection in multispectral document images. This method outperforms existing techniques and shows promise for forensic analysis in various fields, including computational forensics.

Hedjam and Cheriet \cite{hedjam2012hyperspectral} introduce an algorithm for hyperspectral band selection, utilizing graph clustering techniques. Their approach involves constructing a band adjacency graph, where the bands are represented as nodes and the similarity between bands is captured by the edges. By applying Markov clustering, they are able to identify coherent clusters of highly correlated bands. In contrast, Martinez et al. \cite{martinez2007clustering} propose a hierarchical clustering structure that focuses on minimizing specific distance measures (such as KL-divergence and MI) to alleviate redundancy among neighboring bands.

Deep learning approaches have also been applied to hyperspectral document image analysis. Khan et al. \cite{khan2018deep} propose a method that extracts spectral responses of ink pixels from hyperspectral images using convolutional neural networks (CNN). This approach achieves high accuracy in ink mismatch detection, surpassing existing techniques. Combining deep learning with spectral correlation and spatial context \cite{khan2019spatio} presents a method for ink mismatch detection and document authentication. The proposed approach demonstrates effectiveness in distinguishing visually similar inks and holds promise for enhancing document authentication techniques using hyperspectral imaging. Furthermore, writer identification is a crucial task in forensic document analysis \cite{islam2019hyperspectral}, where a combined hyperspectral imaging and deep learning approach improves accuracy and robustness in writer identification, highlighting the potential of hyperspectral imaging and deep learning for reliable identification in forensic applications.

To facilitate research and benchmarking, the availability of datasets is crucial. Islam et al. \cite{islam2022ivision} introduce a dataset of hyperspectral images of handwriting samples. This dataset is a valuable resource for evaluating and advancing algorithms and methods in tasks such as ink mismatch detection, writer identification, and forgery detection.

Hyperspectral imaging has shown tremendous potential in forensic document analysis, enabling improved ink mismatch detection, forgery detection, writer identification, and document authentication. The reviewed literature demonstrates the effectiveness of hyperspectral imaging in overcoming the limitations of traditional RGB scans and existing techniques. Further research and advancements in sensor technology, spectral analysis methods, and deep learning approaches will undoubtedly enhance the capabilities of hyperspectral imaging and contribute to forensic document analysis.

\section{Methodology}
\begin{figure}[htbp]
\centerline{\includegraphics[scale=0.35]{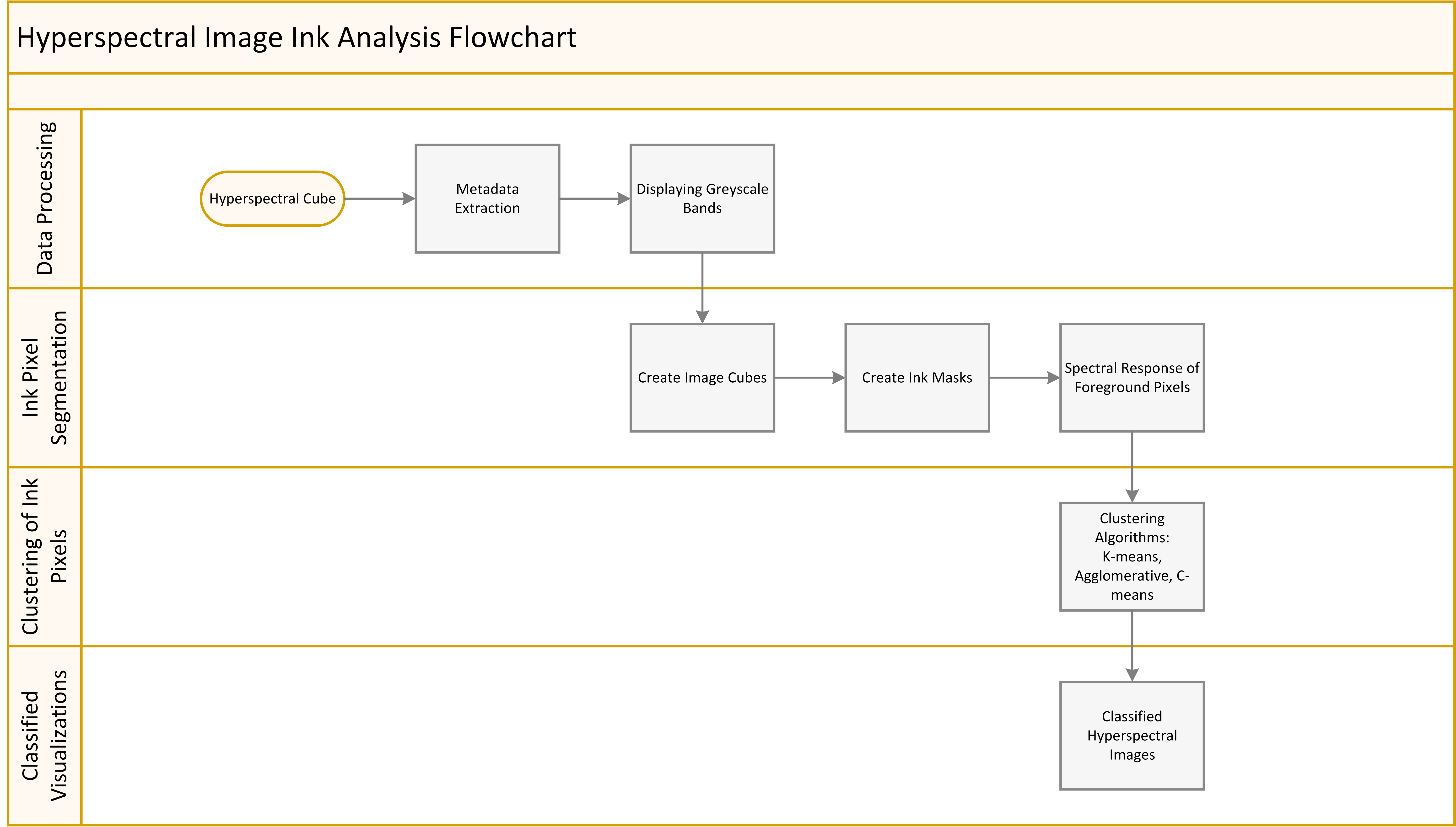}}
\caption{Hyperspectral Image Ink Analysis Flowchart}
\label{fig: Workflow}
\end{figure}

This section outlines the scientific methodology employed in this research study to analyze a hyperspectral cube dataset obtained from the iVision HDD dataset. The workflow (Fig.~\ref{fig: Workflow}) encompassed various steps, including extraction of data information, segmentation of ink pixels, and determination of the number of inks present in the document.

\subsection{Data Processing}
The first step was the extraction of metadata from the header file, providing crucial details (as shown in table 1) for subsequent analysis. The hyperspectral cube obtained from the iVision HDD dataset was a handwritten document with a resolution of 512 X 650 pixels, with a total of 149 bands. The spectral range spanned from 478 nm to 900 nm. Each hyperspectral band was then visualized as grayscale images, allowing for a detailed examination of spectral characteristics.

\begin{table}[htbp]
\caption{Details of HSI Cube}
\begin{center}
\begin{tabular}{|c|c|c|c|}
\hline
Rows & 650\\
\hline
Samples & 512 \\
\hline
Bands & 149\\
\hline
Interleave & BSQ\\
\hline
Quantization & 32 bits \\
\hline
Data Format & float32 \\
\hline
Starting Wavelength & 478.7825462 nm \\
\hline
Ending Wavelength & 900.9723394 nm \\
\hline
\end{tabular}
\label{tab1}
\end{center}
\end{table}

\subsection{Ink Pixel Segmentation}
The next step was to separate the handwritten text from the blank paper area. The HSI cube was cropped in order to focus exclusively on the area of interest and eliminate extraneous information, isolating the foreground's spectral response and removing irrelevant data from the sides. 

The resulting cropped image cube was converted to grayscale, enabling the analysis of pixel intensities in a monochromatic representation. A line curve was also plotted to visualize the distribution of reflectance values for mean ink pixels across the image. Line cropping was performed using a binary image generated in the previous step to isolate ink pixels within each line. Boolean masks were created for each line, selectively retaining only the ink pixels. This process resulted in line-specific images containing exclusively the ink pixels.

The line-specific ink pixel extraction produced 2D images, which were then converted back into hyperspectral cubes, generating a set of 12 cubes representing different lines. Spectral responses were plotted for the ink pixels in each line, providing insights into the variations in spectral characteristics across different lines.

Within each hyperspectral cube representing a line, ink pixels were extracted by applying a Boolean mask, and the mean value of the ink pixels within each line was computed using a built-in mean function. This approach efficiently segmented ink pixels within the document. Utilizing the mean values obtained from each line, the spectral responses of all 12 lines were plotted, enabling a visual representation of the spectral characteristics exhibited by the ink pixels across different lines.

To determine the number of inks, present in the document, a Boolean mask encompassing the entire binary image was generated using the same methodology employed for line-specific masks. The image was inverted to transform black ink pixels into white, while the background remained black.

\subsection{Clustering of Ink Pixels}
Three clustering algorithms, including k-means, agglomerative, and c-means, were applied to the inverted image to create clusters of inks. The k-means clustering algorithm, with a specified number of clusters set to 7, iteratively reassigned pixels and resulted in 7 distinct clusters representing different ink colors. Similarly, agglomerative and c-means clustering was performed, resulting in 7 different ink colors.

The clusters generated by the clustering algorithms were analyzed to identify the distinct ink colors present in the document, providing insights into the clustering patterns and facilitating the determination of the number of inks based on the resulting clusters. Finally, the resultant classifications are classified based on the ink clustered and visualized. 

In summary, this methodology employed a systematic workflow that encompassed data extraction, ink pixel characterization, and ink number determination. By combining image processing techniques and clustering algorithms, a comprehensive analysis of the hyperspectral cube dataset was conducted, revealing valuable insights into the spectral behavior and ink composition within the document.

\section{Results and Discussions}
\begin{figure}[htbp]
\centerline{\includegraphics[scale=0.4]{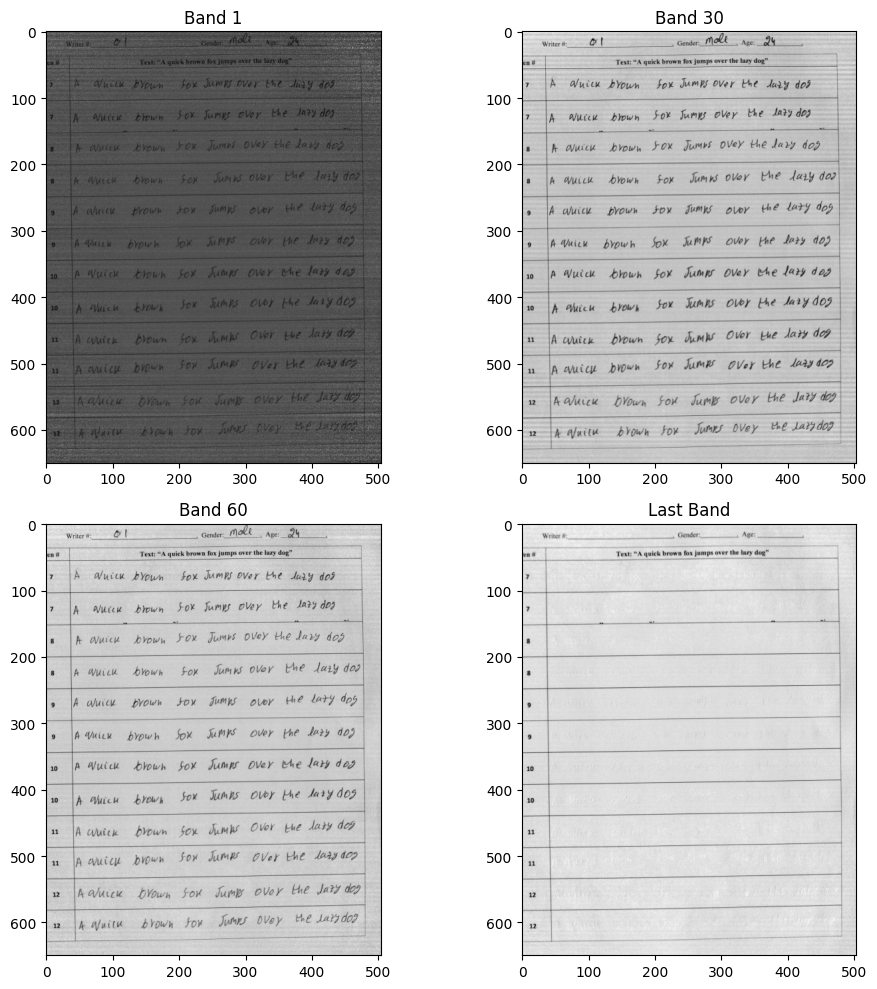}}
\caption{Showing different bands of HSI (1\textsuperscript{st}, 30\textsuperscript{th}, 60\textsuperscript{th} and 149\textsuperscript{th})}
\label{fig: F1}
\end{figure}

In this section, we will discuss the findings derived from analyzing a dataset consisting of a hyperspectral cube. To illustrate our observations, (Fig.~\ref{fig: F1}) displays a selection of bands from the hyperspectral images presented as individual grayscale images. Upon examining these images, it becomes apparent that the initial bands exhibit lower levels of spectral reflectance, resulting in a greater number of grey shades. However, as the wavelength increases, the spectral responses improve, leading to clearer text. Around band 60, all 12 lines of text are quite legible. As the wavelength continues to increase, the text gradually fades, ultimately disappearing entirely by band 98.

\begin{figure}[htbp]
\centerline{\includegraphics[scale=0.45]{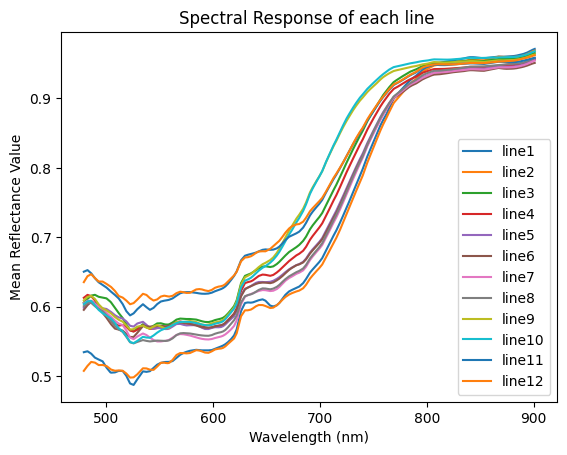}}
\caption{Spectral response of 12 lines}
\label{fig: F2}
\end{figure}

Comparing these selected bands with their corresponding spectral responses (Fig.~\ref{fig: F2}), it is apparent that the mean reflectance value gradually increases in the optical wavelength region. However, as the wavelength extends into the near-infrared region, the spectral response stabilizes. Additionally, Fig.~\ref{fig: F2} demonstrates that the 12 lines exhibit different spectral responses in the optical wavelength range, indicating the presence of different inks in the document. However, as the wavelength increases, the spectral responses become more similar. Analyzing the spectral responses, it was determined that the hyperspectral image contained 7 different inks.

\begin{figure}[!htb]
\minipage{0.3\textwidth}
  \includegraphics[width=\linewidth]{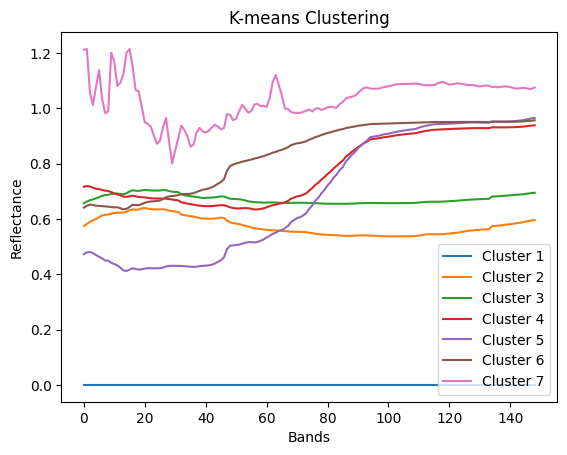}
\endminipage\hfill
\minipage{0.3\textwidth}
  \includegraphics[width=\linewidth]{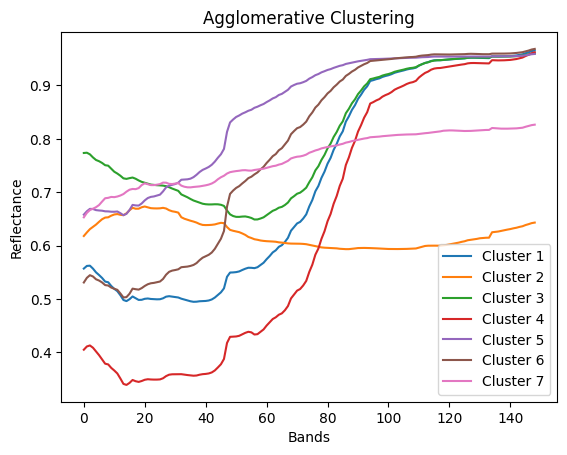}
\endminipage\hfill
\minipage{0.3\textwidth}%
  \includegraphics[width=\linewidth]{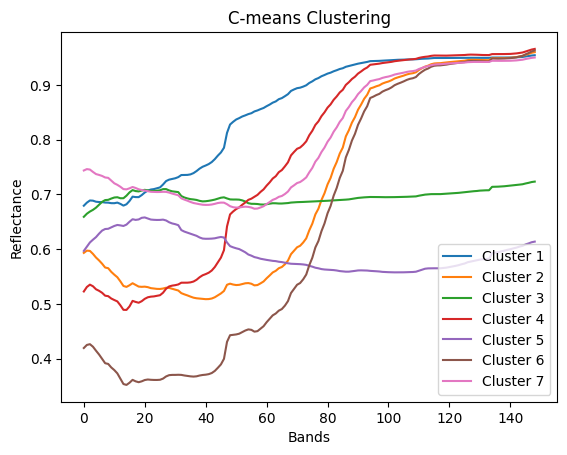}
\endminipage
\caption{The spectral responses of different ink clusters using (a) k-means Clustering, (b) Agglomerative Clustering and (c) c-means Clustering}
\label{fig: F3}
\end{figure}

To identify the ink clusters, three clustering approaches, namely k-means clustering, c-means clustering, and Agglomerative clustering, were employed. Fig.~\ref{fig: F3} displays the results of these clustering techniques, showcasing somewhat similar patterns. k-means clustering identified two distinct ink clusters: ink cluster 1, representing background pixels with zero reflectance across all bands, and ink cluster 7, displaying a unique pattern compared to other inks. On the other hand, Agglomerative clustering and C-means clustering produced similar results, without exhibiting any distinct behavior among the clusters.

\begin{figure}[!htb]
\minipage{0.20\textwidth}
  \includegraphics[width=\linewidth]{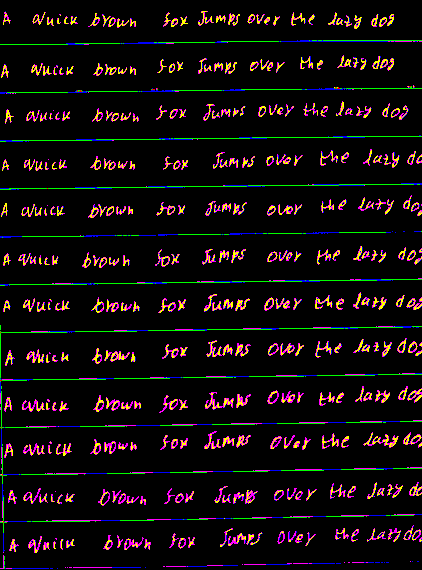}
\endminipage\hfill
\minipage{0.25\textwidth}
  \includegraphics[width=\linewidth]{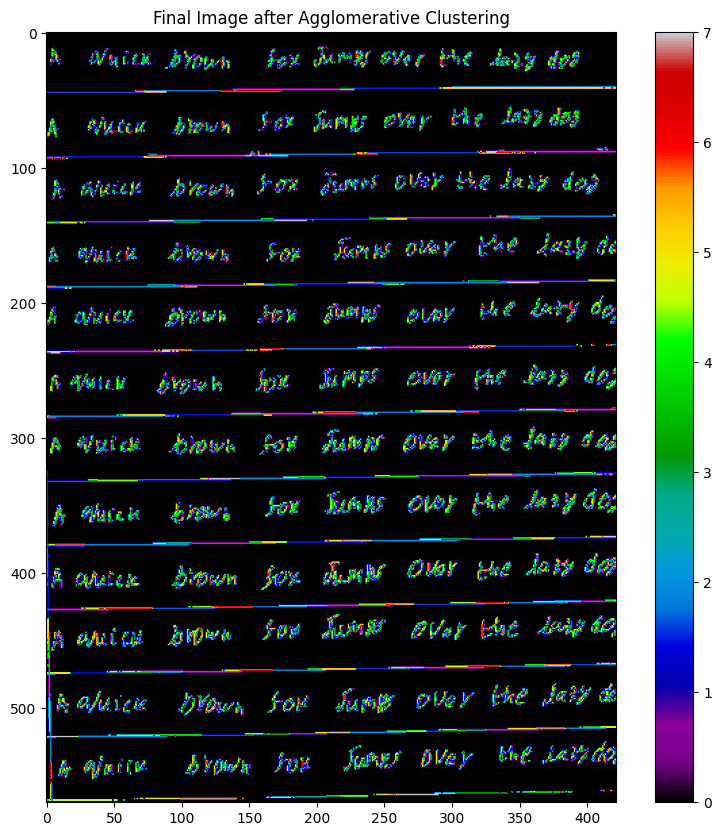}
\endminipage\hfill
\minipage{0.25\textwidth}%
  \includegraphics[width=\linewidth]{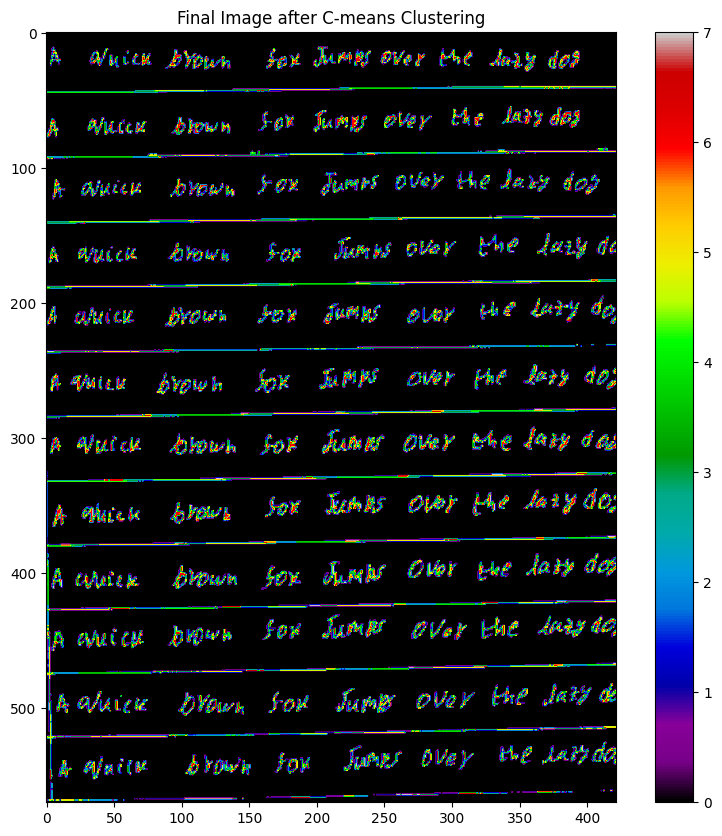}
\endminipage
\caption{Classified Hyperspectral Images: (a) k-means Clustering, (b) Agglomerative Clustering and (c) c-means Clustering}
\label{fig: F4}
\end{figure}

Fig.~\ref{fig: F4} presents the final classified hyperspectral images, highlighting the different ink clusters. The results from k-means clustering appear to classify the inks more effectively compared to the other two approaches. The outcomes of c-means clustering and Agglomerative clustering exhibit similar patterns.

In conclusion, the analysis of the hyperspectral cube dataset revealed interesting insights. The spectral reflectance varied across different bands, with clearer text observed in higher wavelengths. The optical wavelength region exhibited diverse spectral responses among the 12 lines, indicating the presence of multiple inks. Clustering algorithms successfully identified ink clusters, with k-means clustering demonstrating better classification performance. These findings contribute to a deeper understanding of the spectral behavior and ink composition within the document, paving the way for further investigations in the field of hyperspectral imaging and ink analysis.

\section{Conclusion}
In conclusion, this study demonstrated the potential of hyperspectral imaging in enhancing the scientific precision of ink analysis. By utilizing techniques like clustering algorithms, the proposed methodology successfully identified the number of distinct inks present in a document. The findings contribute to a deeper understanding of spectral behavior and ink composition, thereby safeguarding the authenticity and credibility of legal records and historical artifacts.

Future works in ink analysis using hyperspectral imaging should focus on several areas. Firstly, further advancements in sensor technology and spectral analysis methods can enhance the capabilities of hyperspectral imaging, allowing for more accurate ink discrimination and quantitative analysis. Additionally, the integration of deep learning approaches with hyperspectral imaging can provide more robust and automated techniques for ink mismatch detection, forgery detection, and writer identification. Furthermore, the development of standardized datasets and benchmarks will facilitate research and comparison of different methodologies in the field.

\bibliographystyle{IEEEtran}
\bibliography{references}

\vspace{12pt}

\end{document}